\documentclass[10pt, a4paper]{article}

\usepackage{lrec-coling2024} 

\usepackage[T1]{fontenc}
\usepackage{bm}
\usepackage{graphicx}
\usepackage{makecell}
\usepackage{color}
\usepackage{multirow}
\usepackage{tabularx}
\usepackage{stfloats}
\usepackage{cprotect}
\usepackage{amsmath}
\usepackage{setspace}
\usepackage{hyperref}
\usepackage{cleveref}
\usepackage{ulem}
\usepackage{verbatimbox}
\usepackage{booktabs}
\usepackage{svg}

\definecolor{myGreen}{rgb}{0.76,0.93,0.63}
\definecolor{myRed}{rgb}{0.93,0.5,0.5}
\definecolor{LimeGreen}{rgb}{0.2,0.8,0.2}

\title{Improving Factual Error Correction for Abstractive Summarization via Data Distillation and Conditional-generation Cloze}

\name{\makecell{Yiyang Li$^{*1}$, Lei Li$\;^{\dagger*1}$, Dingxin Hu$^{1}$, Xueyi Hao$^{1}$, 
\\ Marina Litvak$^{2}$, Natalia Vanetik$^{2}$, Yanquan Zhou$\;^{\dagger1}$}} 

\address{$^1$Beijing University of Posts and Telecommunications \\
         $^2$Shamoon College of Engineering \\
         \{kenlee, leili, zhouyanquan\}@bupt.edu.cn}

\abstract{
Improving factual consistency in abstractive summarization has been a focus of current research. One promising approach is the post-editing method. However, previous works have yet to make sufficient use of factual factors in summaries and suffers from the negative effect of the training datasets. In this paper, we first propose a novel factual error correction model FactCloze based on a conditional-generation cloze task. FactCloze can construct the causality among factual factors while being able to determine whether the blank can be answered or not. Then, we propose a data distillation method to generate a more faithful summarization dataset SummDSC via multiple-dimensional evaluation. We experimentally validate the effectiveness of our approach, which leads to an improvement in multiple factual consistency metrics compared to baselines.
 \\ \newline  }

\begin{document}
\maketitleabstract

\renewcommand{\thefootnote}{\fnsymbol{footnote}}
\footnotetext[1]{These authors contributed equally.}
\footnotetext[2]{Corresponding authors.}
\renewcommand{\thefootnote}{\arabic{footnote}}

\section{Introduction}
\label{sec:Introduction}
In recent years, abstractive summarization has achieved great progress based on the development of deep learning and pre-trained language models.  However, a number of works\citep{cao2018faithful, maynez2020faithfulness, deutsch2021understanding} have shown that SOTA models still suffer from factual inconsistencies. This problem hinders the application of abstractive summarization. To improve the faithfulness of the summaries, recent works focus on the post-editing methods. It is a plug-and-play approach that corrects the factual errors in the summaries generated by the summarization models. 

\begin{table}[t]
\renewcommand\arraystretch{1.1}
\centering
\scalebox{0.85}{
\begin{tabular}{p{0.25\textwidth}|p{0.25\textwidth}}
\toprule[0.5mm]
\textbf{\makecell[c]{Cold-boot Method}} & \textbf{\makecell[c]{Warm-boot Method}}\\
\hline
\multicolumn{2}{c}{\textbf{Training} $\{(d,s)\}$} \\
\hline
                              & {\textbf{\makecell[c]{Data Augmentation}}} \\
 \makecell[c]{\textbf{Cloze/QA/Other Task}} & \makecell[c]{$s + noise \to s^-$} \\
                                \cline{2-2}
\makecell[c]{\textbf{with Public Datasets}} & {\textbf{\makecell[c]{Conditional Generation}}} \\
                              & \makecell[c]{$d + s^- \stackrel{M}{\to} s$} \\
\hline
\multicolumn{2}{c}{\textbf{Inference} $\{(d,h)\}$} \\
\hline
\textbf{\makecell[c]{Extract Factual Factors}} & \\ 
\makecell[c]{$h\stackrel{E}{\to}\{f_{1},..,f_{k},...,f_{K}\}$} & \\
\cline{1-1}
\textbf{\makecell[c]{Obtain Candidates}} & \textbf{\makecell[c]{Correct Factual Errors}} \\ 
\makecell[c]{$d\stackrel{E}{\to}\{c_{1},..,c_{k},...,c_{K}\}$} & \textbf{\makecell[c]{$d+h\stackrel{M}{\to}h^{\prime}$}} \\ 
\cline{1-1}
\textbf{\makecell[c]{Correct Factual Errors}} & \\
\makecell[c]{$Substitute(c_{k},f_{k})\to{h^{\prime}}$} & \\

\hline
\end{tabular}}
\caption{Comparison between cold-boot and warm-boot methods through a formulaic form. \textbf{Cold-boot method}: During the training phase, it builds cloze, question answering (QA) or other tasks and trains an extractor $E$ with public datasets. In the inference phase, $E$ is used to extract factual factors from the document $d$ and hypothesis $h$. The correction progress is treated by removing the original factual factors ($f_{k}$) in $h$ and substituting the ones ($c_{k}$) in $d$. \textbf{Warm-boot method}: This method is required to construct the training dataset in the first place, which includes the document $d$, an unfaithful summary $s^-$, and a corrected summary $s$. Then, an end-to-end error correction model $M$ is trained on this dataset. The input is the concatenation of document $d$ and the hypothesis $h$ while the output is the corrected summary $h^{\prime}$.}
\label{tab:1}
\end{table}

As shown in Table \ref{tab:1}, we classify the existing works into two categories of the warm-boot methods \citep{cao2020factual,balachandran2022correcting,fabbri2022improving} and the cold-boot methods \citep{dong2020multi,chen2021improving,lee2022factual} through a formulaic form. The warm-boot methods consider factual error correction as a text generation task, where the concatenation of the document and the model-generated summary (hypothesis) is the input and the corrected summary is the output. These methods rely heavily on the construction of positive and negative samples, and most researchers focus on how to generate datasets that are similar to the real distribution. 

In contrast, the cold-boot methods pay more attention to extracting factual factors. They introduce the cloze-based, QA-based, and other tasks to extract the factual factors\footnote{The factual factors denote the text spans that describe facts, such as entities, noun phrases, etc.} from the document and substitute them for the incorrect factual factors in the hypothesis. Thus, they correct the factual factors one by one and independently, ignoring the causality among them. Meanwhile, they cannot explicitly predict which factual factors need to be corrected so that all factual factors in the hypothesis will be replaced. We define these two problems as "Independent Correction Problem" and "Over-Correction Problem", which not only decrease correction efficiency but also introduce new factual errors. 

Faced with the pros and cons of the two categories of methods, we introduce a cloze model via conditional generation task and propose a factual error correction model FactCloze in cold-boot framework. We first mask the factual factors in the hypothesis and input it with the document to FactCloze. The corrected summary is generated by filling in the masked spans. Inspired by the warm-boot methods, we also propose a data distillation method to generate a highly faithful dataset SummDSC to train FactCloze.

Our main contributions are as follows. First, we propose a novel factual error correction model FactCloze based on a conditional-generation cloze task, which solves the "Independent Correction Problem" and "Over-Correction Problem" in cold-boot methods. Second, we construct a highly faithful dataset SummDSC for FactCloze through multi-dimensional data distillation and analyze its plausibility. Third, we validate the effectiveness of FactCloze and SummDSC through a series of experiments on public datasets, where we achieve the best performance compared to strong baselines. 
Codes and models are released at \href{https://github.com/Mr-KenLee/FactCloze}{https://github.com/Mr-KenLee/FactCloze}.

\section{Related Work}
\subsection{Factual Consistency Metrics}
\paragraph{NLI-based metrics} \citet{falke2019ranking}, \citet{barrantes2020adversarial} and \citet{kryscinski2020evaluating} train a natural language inference (NLI) model and evaluate the factual consistency via the entailment score between the document and the hypothesis. \citet{laban2022summac} divide a document and a hypothesis into multiple blocks and apply the NLI model to calculate the blocks for an entailment matrix to resolve the problem of the granularity mismatch.

\paragraph{Dependency-based metrics} \citet{goyal2020evaluating} propose a dependency arc entailment (DAE) method and assign factual consistency scores by comparing whether each dependency arc is entailed by the document. \citet{goyal2021annotating} conduct further studies on data augmentation and model structure, resulting in improvements to the DAE.

\paragraph{QA-based metrics} \citet{wang2020asking}, \citet{durmus2020feqa} and \citet{scialom2021questeval} employ a two-stage approach to evaluate factual consistency. First, they generate questions using a question generation module and ask both the document and the hypothesis. The final factual consistency score is determined by comparing the similarity of the answers given by the document and the hypothesis to the same question. \citet{fabbri2021qafacteval} conduct experiments and optimize each module of the QA-based methods, resulting in better performance compared to other QA-based metrics.

\paragraph{Cloze-based metrics} \citet{li2022just} propose a cloze-based framework for evaluating factual consistency. By masking factual factors in the hypothesis, they generate cloze questions and use a cloze model to obtain answers. The final evaluation score is determined by comparing the similarity between the masked factual factors and the cloze answers.

\subsection{Factual Error Correction}
\paragraph{Cold-boot methods} 
\citet{dong2020multi} and \citet{lee2022factual} use a single model for retrieving correct factual factors from the document to correct the hypothesis. 
\citet{chen2021improving} introduce a factual consistency metric to assist in determining the validity of current error corrections.
\citet{dong2022faithful} further introduce a knowledge graph to address the problem of extrinsic hallucinations.

\paragraph{Warm-boot methods} 
\citet{cao2020factual} and \citet{zhu2021enhancing} follow the approach of \citet{kryscinski2020evaluating} to generate training pairs for a Seq2Seq-based factual error correction model.
\citet{balachandran2022correcting} enhance the data augmentation approach by employing a cloze model to generate negative samples containing factual errors. \citet{fabbri2022improving} introduce factual errors into faithful summaries through a text compression task.

\subsection{Faithful Summarization Datasets}
\label{sec:Faithful Summarization Datasets}
The faithfulness of summarization datasets has recently gained much attention. Researchers such as \citet{guo2022questioning}, \citet{nan2021entity}, \citet{aharoni2022mface}, \citet{choubey2021cape}, \citet{wan2022factpegasus} have constructed more reliable summarization datasets by evaluating and removing samples with low factual consistency scores using one or more factual consistency metrics.

\section{\textbf{FactCloze}}

\begin{figure}
\includegraphics[width=210pt]{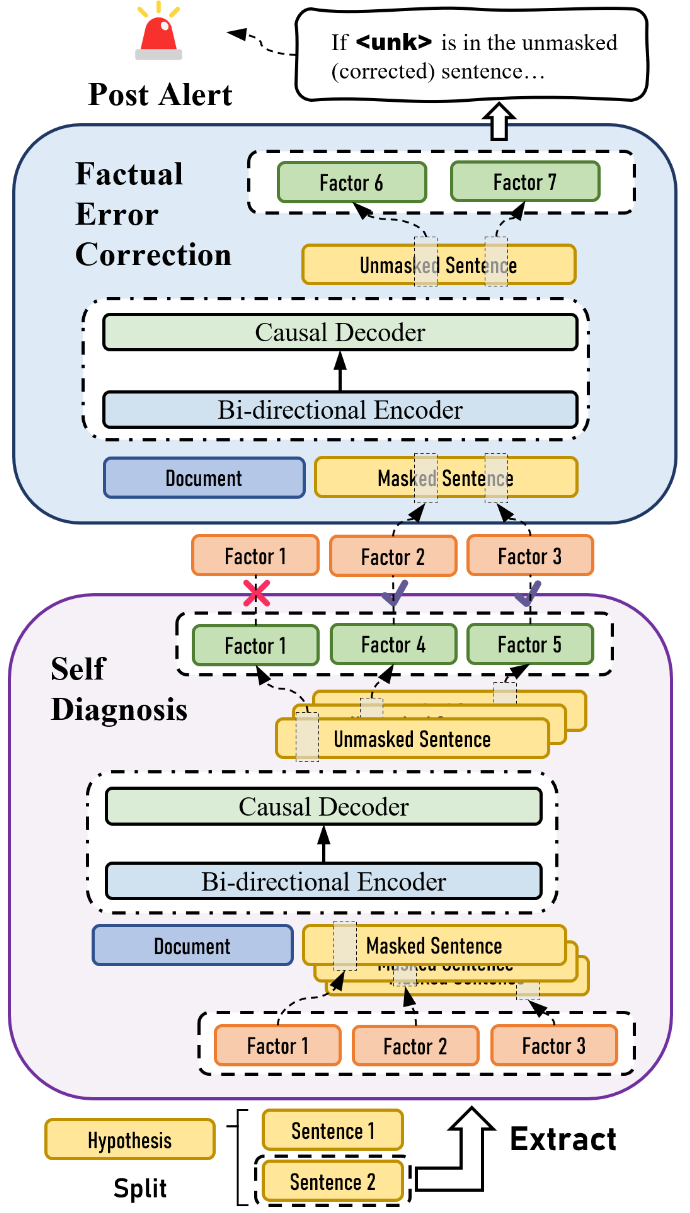}
\cprotect\caption{Overview of \textbf{FactCloze}. A hypothesis sentence is passed to a self-diagnosis mechanism and a factual error correction module. An alert will be raised if the corrected sentence contains \verb|<unk>|s. }
\label{figure:1}
\end{figure}

Given a document $\bm{d}=\{d_1,...,d_N\}$ and a hypothesis $\bm{h}=\{h_1,...,h_M\}$, the task of factual error correction model $M_c$ is to generate a corrected summary $\bm{h}^\prime=\{h_1^\prime,...,h_L^\prime\}$. Each token $d_n$, $h_m$ and $h_l^\prime$ takes one value from a vocabulary $\mathcal{V}$.
Formally, the generation probability of $\bm{h}^\prime$ is formulated as: 
\begin{equation}
\label{equation:1}
P(\bm{h}^\prime|\bm{d},\bm{h})=\prod_{l=1}^{L}P(h_l^\prime|h_{<l}^\prime,\bm{d},\bm{h})
\end{equation}
As described in \S\ref{sec:Introduction}, we can improve faithfulness by directly correcting the factual factors $\bm{f}=\{f_1,...,f_K\}$. Thus, the task of factual error correction can be reformulated as: 
\begin{equation}
\label{equation:2}
{P(\bm{h}^\prime|\bm{d},\bm{h})\Rightarrow{P(\bm{f}^{(\bm{h}^\prime)}|\bm{d},\bm{h}_{/\bm{f}^{(h)}})}}
\end{equation}
where $\bm{h}_{/\bm{f}^{(h)}}$ denotes the hypothesis without the factual factors $\bm{f}^{(h)}$ in $\bm{h}$ and $\bm{f}^{(\bm{h}^\prime)}$ denotes the correct factual factors generated by $M_c$. The number of $\bm{f}^{(h)}$ and $\bm{f}^{(\bm{h}^\prime)}$ are both $K$. 

Previous works leverage many kinds of tasks to generate $\bm{f}^{(\bm{h}^\prime)}$ with Eq.\ref{equation:2}. However, they still face two serious problems (\S\ref{sec:Introduction}) that hinder $M_c$ to correct the summary by understanding contextual semantics, which results in low error correction accuracy and efficiency. 

To solve these problems, we introduce the cloze task \citep{taylor1953cloze} to model the correct factual factors generation task and propose a novel factual error correction model FactCloze at the sentence level, as shown in Figure \ref{figure:1}. We consider the concatenation of document $\bm{d}$ and the masked hypothesis $\bm{h}_{/\bm{f}^{(h)}}$ as input and the corrected factual factors $\bm{f}^{(\bm{h}^\prime)}$ as output via filling the masked spans by the cloze model. The corrected summary can be obtained by filling the $\bm{f}^{(\bm{h}^\prime)}$ into the $\bm{h}_{/\bm{f}^{(h)}}$.

\subsection{Cloze Model}
Due to the similarity between the cloze task and the masked language modeling task (MLM) which is applied in pre-trained language models (PLMs) widely, we can adopt any PLM that has been trained on MLM task as a cloze model. 
However, the autoencoder-like PLMs \citep{kenton2019bert,liu2019roberta} cannot causally model cloze task because of the independence assumption of masked spans generation \citep{yang2019xlnet, du2022glm}. 
Thus, we use BART\citep{lewis2020bart} and T5\citep{raffel2020exploring} as the cloze models. Their autoregressive decoders can naturally model the causality among the factual factors, which solves the "Independent Correction Problem". In this case, Eq.\ref{equation:2} can be further reformulated as:
\begin{align}
\label{equation:3}
{P(\bm{f}^{(\bm{h}^\prime)}|\bm{d},\bm{h}_{/\bm{f}^{(h)}})} \hspace{3cm} \nonumber \\
=\prod^{K}_{k=1}P(f^{(\bm{h}^\prime)}_k|f^{(\bm{h}^\prime)}_{<k}, \bm{d},\bm{h}_{/\bm{f}^{(h)}})
\end{align}
where the equal sign holds only in the autoregressive generation mode.

\subsection{Training}
We train BART and T5 based on their corresponding MLM tasks separately. For a document and its faithful summary, we randomly select several factual factors (i.e. named entities and noun phrases) in the summary and mask them with an equal number of \verb|[MASK]| tokens. The encoder inputs for both models are a concatenation of the document and the masked summary. The goal for T5 is to generate factual factors for each \verb|[MASK]| position, while BART aims to generate the unmasked summary. We use teacher forcing \citep{williams1989learning} and a cross-entropy loss for both BART and T5.

\subsection{Inference}
At inference time, we use beam search \citep{sutskever2014sequence} to decode the generation. Due to the differences in generation between BART and T5, we employ different strategies to obtain the final corrected summary. BART generates the corrected summary directly with its decoder, while T5 generates cloze answers for the masked spans and merges these answers with the masked hypothesis to obtain the corrected summary. Moreover, we propose a self-diagnosis mechanism to alleviate the “Over-Correction Problem”.
\paragraph{Self Diagnosis}
\label{sec:Self Diagnosis}
The “Over-Correction Problem” occurs when all factual factors $\bm{f}^{(h)}$ in the hypothesis are corrected, even though some of them are faithful. For this reason, we adopt the idea of ClozE \citep{li2022just} to solve this problem and propose a self-diagnosis mechanism.
Firstly, we mask $\bm{f}^{(h)}$ and answer the masked spans by the cloze model one by one. Afterward, we discard the factual factors whose answers are consistent with the original ones and obtain a subset as follows:
\begin{equation}
\label{equation:4}
\bm{f}^{(c)}=\{f^{(c)}_i|f^{(c)}_i\neq{f^{(c^\prime)}_i}, i=1,2,...,K\}
\end{equation}
where $f^{(c^\prime)}_i\sim{P(f^{(c^\prime)}_i|\bm{d},\bm{h}_{/f^{(h)}_i})}$. Finally, all the remaining factual factors $\bm{f}^{(c)}$ are masked and answered at once to obtain the corrected summary.

\subsection{Post Alert}
\label{sec:Post Alert}
Using a factual error correction model to correct all hypotheses may result in lower faithfulness and unknown risks. This is because not all hypotheses can be corrected by the correction model \citep{chen2021improving,pagnoni2021understanding}, such as the text which is completely irrelevant to the document. In practice, it is necessary to identify these risky hypotheses. To address this issue, we propose a post-alert mechanism that allows FactCloze to determine whether a hypothesis can improve its faithfulness by correction. We introduce the \verb|<unk>| token as a special factual factor. If FactCloze fills in \verb|<unk>|s for masked spans, it will raise an alert. We enable this capability in FactCloze by constructing a specific training dataset, as described in \S\ref{sec:Alert Support}.

\section{SummDSC}
\label{sec:SummDSC}
\begin{figure}
\includegraphics[width=220pt]{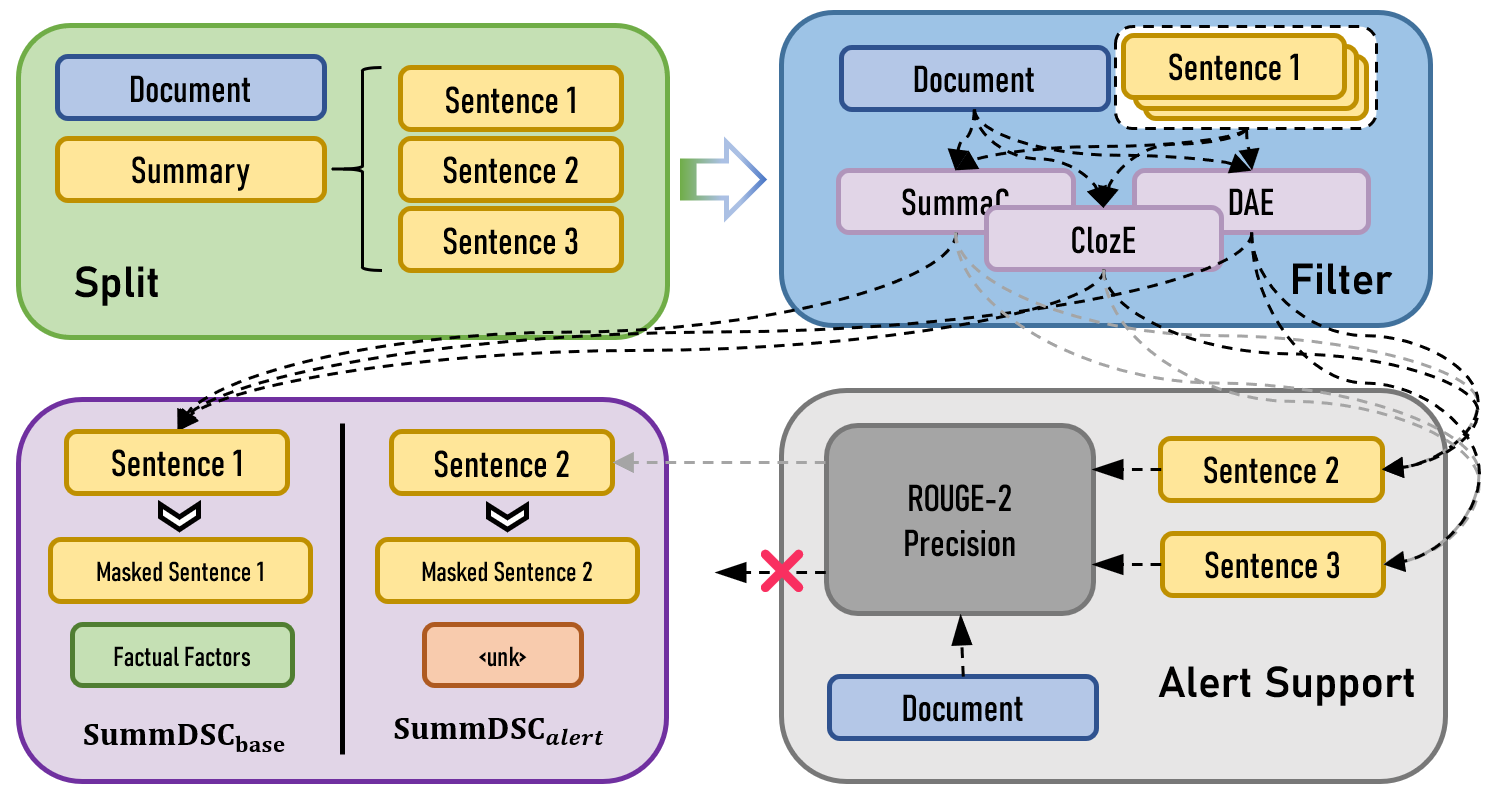}
\caption{Overview of SummDSC. We use four modules to convert a document-summary pair to $\text{SummDSC}_{base}$ and $\text{SummDSC}_{alert}$ formats. The black dashed line indicates that the factual consistency score is above the threshold, while the opposite is true for the gray ones.}
\label{figure:2}
\end{figure}

We construct the training dataset for FactCloze using the public summarization datasets CNN/DM \citep{hermann2015teaching} and XSum \citep{narayan2018don}. However, these datasets suffer from the unfaithful problem (\S\ref{sec:Faithful Summarization Datasets}), which significantly impacts the accuracy of the correction models trained on them. To address this issue, we propose a data distillation method to generate SummDSC dataset for FactCloze, as shown in Figure \ref{figure:2}. Specifically, SummDSC can be further split into two subsets, i.e. faithful summarization dataset $\text{SummDSC}_{base}$ and post-alert dataset $\text{SummDSC}_{alert}$.

\subsection{Multi-dimensional Filtering}
\label{sec:Multi-dimensional Filtering}
Previous work \citep{pagnoni2021understanding, tang2022understanding} has shown that different categories of metrics are not equally sensitive to different factual error types. Thus, we develop previous filtering strategies \citep{nan2021entity, guo2022questioning} and 
filter the datasets using the factual consistency metrics in three dimensions, which are dependency, NLI and QA. We take into account the performance and evaluation speed and choose DAE \citep{goyal2020evaluating}, SummaC \citep{laban2022summac} and ClozE\footnote{We adopt ClozE as a substitute for QA-based metrics due to its similar evaluation process.} \citep{li2022just} to guide the filtering process. Each metric has been set a threshold, as described in Appendix A. Any datapoint with a factual score lower than the corresponding threshold will be discarded and the remaining ones form the filtered dataset $\text{SummDSC}_{base}$.

\subsection{Alert Support}
\label{sec:Alert Support}
To support the post-alert mechanism, we generate $\text{SummDSC}_{alert}$ with the discarded datapoints in \S\ref{sec:Multi-dimensional Filtering}. Because the risky hypothesis is usually full of extrinsic hallucinations, we leverage ROUGE-2 precision \citep{lin2004rouge} and set a threshold (Appendix \ref{sec:Threshold Selection}) to select the post-alert samples from the discarded datapoints. Samples with scores lower than the threshold are included in $\text{SummDSC}_{alert}$. The factual factors in the summaries will be replaced with \verb|<unk>|s in $\text{SummDSC}_{alert}$.

\section{Experiments}
In this section, we first present the experimental settings and implementation details. Then, we conduct experiments to demonstrate the plausibility of our multi-dimensional filtering method. Third, we show the performance of both FactCloze and SummDSC through a series of experiments. Finally, we conduct a manual evaluation to provide a more accurate assessment of our methods.


\subsection{Experimental Settings}
\label{sec:Experimental Settings}
\paragraph{Benchmark Dataset} We firstly experiment on the FRANK \citep{pagnoni2021understanding} dataset, which contains summaries generated by different models on CNN/DM and XSum datasets. We split all test samples into document-sentence pairs, which results in 3915 test sample pairs on CNN/DM and 1027 test sample pairs on XSum.
Moreover, we further experiment on the summaries generated by a BART model on full CNN/DM and XSum datasets, which is most widely used in previous work. 

\paragraph{Automatic Evaluation} We mainly use factual consistency metrics to evaluate the results of factual error correction models. In addition to the DAE, SummaC and ClozE employed in data distillation, we also introduce QAFactEval \citep{fabbri2021qafacteval} and FactCC \citep{kryscinski2020evaluating} in order to make an evaluation more objective. We also consider the ROUGE-2 (F1) metric as a reference for informativeness.

\paragraph{Baselines} We use SpanFact \citep{dong2020multi}, BART-FC \citep{cao2020factual}, CogComp \citep{chen2021improving}, FactEdit \citep{balachandran2022correcting}, and CompEdit \citep{fabbri2022improving} as baselines, where SpanFact and CogComp are cold-boot methods while the rest are warm-boot methods. 
In addition to the different factual error correction models, we also introduce the faithful datasets EntityLevel \citep{nan2021entity} and SummFC \citep{guo2022questioning}. 

\subsection{Implementation Details}
The granularity for correction during both training and inference is at the sentence level. Both of BART and T5 use teacher forcing \citep{williams1989learning} and a cross-entropy loss.
We use two pre-trained models BART-Large and T5-Base with parameters from Huggingface \citep{wolf2019huggingface}. The factual factors are extracted by the \verb|en_core_web_trf| model from Spacy\footnote{\url{https://spacy.io/api}}. We use AdamW \citep{loshchilov2018decoupled} optimizer with learning rate 1e-4 and all models are trained for 5 epochs with batch size 32 on 4 NVIDIA GeForce RTX 3090 GPUs.

\begin{figure}
\includegraphics[width=220pt, height=220pt]{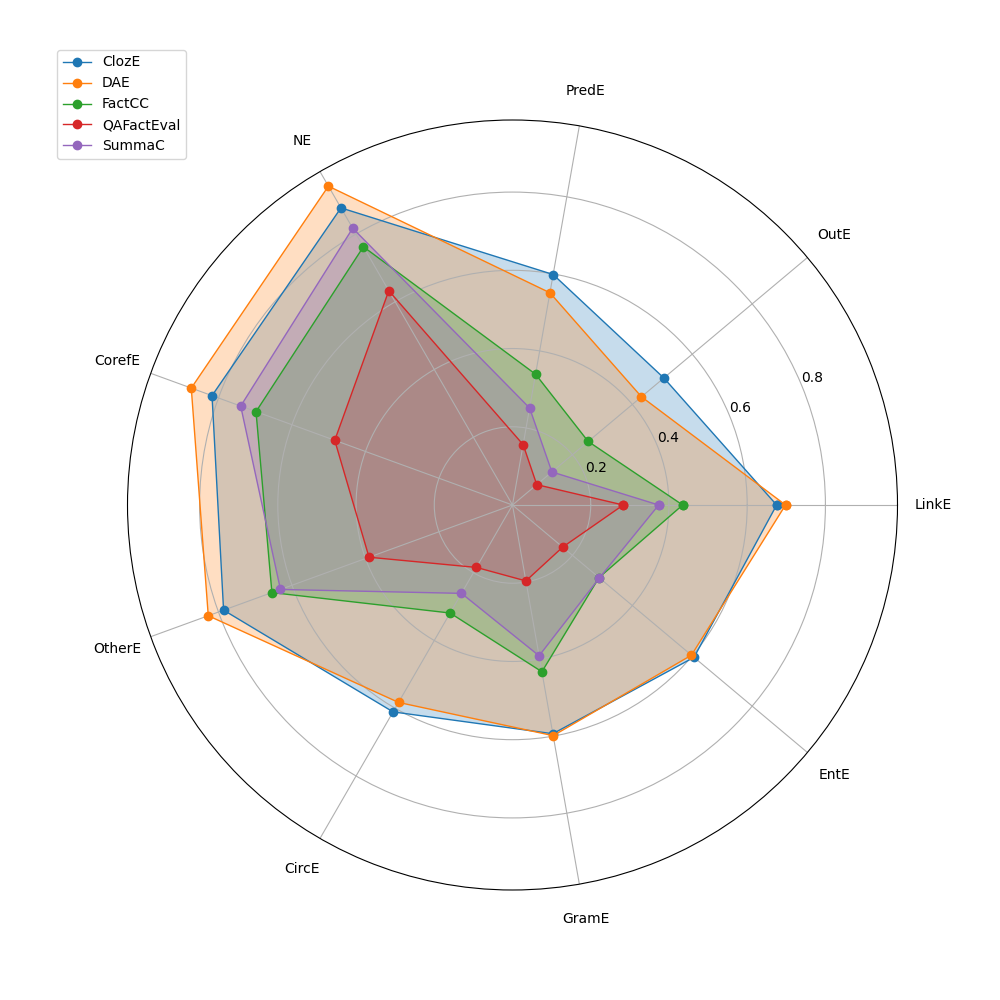}
\caption{A radar chart of the five factual consistency metrics on the FRANK dataset. The different directions indicate the average score on the samples with different error types. The description of each error is referred to \citet{pagnoni2021understanding}. Specially, \textbf{NE} indicates a sample without factual errors.}
\label{figure:3}
\end{figure}

\subsection{Metrics Correlation}
\label{sec:Metrics Correlation}


In our experiment, we also utilize the FRANK dataset to measure the correlation between the five factual consistency metrics in two distinct ways. We first construct a radar chart to display the sensitivity of various metrics to different error types, as shown in Figure \ref{figure:3}. A lower score indicates a higher sensitivity of the metric to the corresponding factual errors. Since different metrics have varying tendencies to score (for example, \textbf{NE} gains a low score via QAFactEval), we analyze them only concerning the metrics themselves. According to the radar chart, various metrics display distinct characteristics and their sensitivity to errors is not uniform. Both DAE and ClozE are more sensitive to predicate errors (\textbf{PredE}), out-of-article errors (\textbf{OutE}), and entity errors (\textbf{EntE}), while FactCC and SummaC also have good recognition of errors such as circumstance errors (\textbf{CircE}). Moreover, QAFactEval generally assigns lower scores in each error type, but it exhibits greater sensitivity to errors in the lower right quadrant of the radar chart.
In another way, we provide a more direct illustration to show the correlation between the different metrics. Following \citet{nan2021improving}, we construct a box plot to show the correlation between factual metrics. The samples are grouped into bins based on the percentiles of one metric score. We then plot the factual consistency score boxes of other metrics within each bin. As shown in Figure \ref{figure:4} in the Appendix, the trends between bins and boxes for different metrics are not consistent enough. It indicates that the correlations between different metrics are not strong, in line with our previous conclusions. Both experiments fully demonstrate the rationality of filtering dataset based on multi-dimensional metrics.

\begin{table*}
\centering
\resizebox{\textwidth}{!}{
\begin{tabular}{clccccc|c}
\hline
\textbf{Type} & \textbf{Methods} &\textbf{QAFactEval(\%)} &\textbf{SummaC(\%)} &\textbf{ClozE(\%)} &\textbf{DAE(\%)} &\textbf{FactCC(\%)} & \textbf{R-2(\%)}\\
\hline
\textbf{Summarization} & Mixtures (FRANK) & 60.01 / 4.72 & 80.66 / 21.08 & 86.73 / 48.14 & 91.25 / 37.91 & 71.25 / 22.16 & 9.76 / 11.79 \\
\hline
\multirow{5}{*}{\textbf{Baselines}} & SpanFact (our imple.) & 56.80 / 5.00 & 77.80 / 21.95 & 84.70 / 48.28 & 91.59 / 38.61 & 66.12 / 22.38 & 9.16 / 11.34\\
                           & BART-FC (our imple.) & 60.93 / 8.60 & 80.54 / 25.99 & 87.71 / 51.06 & 92.98 / 38.89 & 71.34 / 17.41 & 9.32 / 11.27\\
                           & CogComp  & 57.01 / 4.83 & 76.75 / 21.74 & 85.75 / 52.60 & 92.10 / 36.62 & 66.28 / 20.45 & 9.76 / 11.54 \\
                           & FactEdit & 52.95 / 5.09 & 82.04 / 25.63 & 85.25 / 45.91 & 94.49 / 37.26 & 76.82 / 27.98 & 8.90 / 11.20\\
                           & CompEdit & 66.58 / 8.52 & 85.17 / 26.11 & 83.97 / 47.55 & 87.85 / 41.52 & 56.83 / 26.04 & 9.96 / 11.39\\
\hline
                           & BART+Full & 54.72 / 15.86 & 79.20 / 30.02 & 80.28 / 67.09 & 87.32 / 52.52 & 62.46 / 27.70 & 8.77 / 13.02 \\ 
                           & BART+EntityLevel & 55.37 / 15.64 & 79.77 / 31.35 & 84.31 / 68.51 & 86.40 / 53.16 & 61.38 / 27.24 & 8.85 / 13.06 \\
\textbf{FactCloze}         & BART+SummFC & 59.21 / 13.73 & 80.28 / 28.63 & 87.80 / 67.67 & 90.84 / 50.95 & 70.42 / 26.58 & 8.99 / \textbf{13.09} \\
(Different Training Data)  & T5+Full & 64.56 / 8.94 & 84.36 / 24.70 & 93.68 / 67.63 & 95.54 / 49.80 & 73.62 / 25.07 & \textbf{10.22} / 12.50 \\
                           & T5+EntityLevel & 65.39 / 9.20 & 84.53 / 26.98 & 93.64 / 68.56 & 95.76 / 50.49 & 75.83 / 24.16 & 10.19 / 12.18\\
                           & T5+SummFC & 65.35 / 9.11 & 84.49 / 26.61 & 93.87 / 68.54 & 95.85 / 50.25 & 76.55 / 23.94 & 10.14 / 12.11\\
\hline
                           & BART+$\text{SummDSC}_{base}$ & 62.83 / \textbf{16.55} & 83.98 / 31.88 & 89.76 / 68.53 & 93.02 / 54.68 & 72.56 / 28.19 & 9.16 / 12.09 \\ 
\textbf{FactCloze}         & \texttt{ +Self Diagnosis} & 63.24 / 16.32 & 84.05 / \textbf{32.09} & 90.11 / 68.32 & 93.10 / \textbf{56.02} & 73.41 / \textbf{28.20} & 9.26 / 12.18 \\
(Ablation Study)           & T5+$\text{SummDSC}_{base}$ & 65.53 / 8.88 & 84.50 / 26.43  & 93.98 / \textbf{69.91} & 96.12 / 51.85 & 77.72 / 25.14 & 10.12 / 11.81 \\
                           & \texttt{ +Self Diagnosis} & \textbf{66.67} / 8.91  & \textbf{85.36} / 26.56 & \textbf{94.17} / 68.87 & \textbf{96.18} / 51.00 & \textbf{77.87} / 23.90 & 10.11 / 11.99 \\
\hline
\end{tabular}}
\cprotect\caption{Overall results on FRANK dataset, where the left of the cell (*/*) indicates the CNN/DM while the right is XSum. The best performance is marked in bold. The two sections of FactCloze correspond to the effect of different training data and the effect of FactCloze method, respectively.} 
\label{tab:2}
\end{table*}

\subsection{Performance on FRANK}

\subsubsection{FactCloze Performance on FRANK}
Results of applying post-editing models to correct the hypotheses in FRANK are shown in Table \ref{tab:2}. Our results show that correcting factual errors using our model improves factual consistency. Meanwhile, all the models obtain similar informativeness according to the R-2 because factual error correction models will only slightly correct the text spans.

SpanFact and CogComp perform poorly on CNN/DM, and the summaries corrected by it even have a drop in several factual consistency metrics. However, we note that they perform better on the XSum, which is due to the greater number of factual errors on the XSum and the larger correctable space. This means that SpanFact and CogComp cannot provide a good correction for finer-grained errors. BART-FC, FactEdit, and CompEdit perform well in most factual consistency metrics, suggesting that they are effective in improving factual consistency. However, they are not stable for certain metrics. Compared to the original hypotheses in CNN/DM, FactEdit shows a 7.06\% drop in QAFactEval and a 1.48\% drop in ClozE respectively. Similarly, CompEdit also shows a 14.42\% drop in FactCC. For XSum, BART-FC also has a 4.75\% drop in FactCC. These results indicate that they lack generalization and have a preference for factual error correction.

Compared to the baselines, our model achieves the best performance in all factual consistency metrics on both CNN/DM and XSum with strong robustness. We achieve an average improvement of 6.07\% on CNN/DM and 13.51\% on XSum compared to the uncorrected summaries (hypotheses). Comparing FactCloze-BART with FactCloze-T5, we note that FactCloze-T5 is more competitive on CNN/DM while FactCloze-BART performs better on XSum. We believe this is due to the different conditional-generation cloze tasks. FactCloze-BART is directly generating unmasked summaries, which means that it needs to generate the context while answering the masked spans. It exhibits that FactCloze-BART pays more attention to the semantics of the context while answering the masked spans, which is more important in more abstractive summaries, such as XSum. In contrast to the more extractive summary on CNN/DM, understanding the semantics of the context is not as important as that on XSum. In this case, the superiority of the pretrained model becomes more important. This suggests that T5 pretrained by a single cloze task will have better adaptation on this task than BART pretrained by multiple tasks.

\subsubsection{$\text{SummDSC}_{base}$ Performance}
\label{sec:SummDSC Performance}
We experiment with the effect of different faithful summarization datasets for FactCloze, as shown in Table \ref{tab:2}. We train the cloze model in FactCloze using datasets obtained from three filtering methods: Full (no filtering), EntityLevel \citep{nan2021entity}, and SummFC \citep{guo2022questioning}. Overall, all three strategies can train a FactCloze that corrects the hypotheses to some extent, although there is a slight decrease in individual metrics. For instance, FactCloze-BART trained on all three datasets exhibits some degradation on QAFactEval under CNN/DM. In general, EntityLevel and SummFC generally outperform the Full setting. Comparing the effect of EntityLevel and SummFC, SummFC performs better on XSum while EntityLevel is more effective on CNN/DM. However, it appears that neither EntityLevel nor SummFC are quite as effective as FactCloze trained on $\text{SummDSC}_{base}$. The results presented above demonstrate the importance of data filtering and indicate that our multi-dimensional filtering approach performs better. Furthermore, the factual consistency scores achieved by FactCloze with different training datasets are also highly competitive with the baselines, which highlights the robustness of our method.

\begin{table*}
\centering
\resizebox{\textwidth}{!}{
\begin{tabular}{clccccc|c}
\hline
\textbf{Type} & \textbf{Methods} &\textbf{QAFactEval(\%)} &\textbf{SummaC(\%)} &\textbf{ClozE(\%)} &\textbf{DAE(\%)} &\textbf{FactCC(\%)} &\textbf{R-2(\%)}\\
\hline
\textbf{Summarization} & BART-Large                                   & 71.46 / 18.48 & 75.73 / 9.06  & 90.68 / 69.97 & 93.82 / 61.20 & 66.07 / 22.69 & 20.25 / 20.25 \\
\hline
\multirow{3}{*}{\textbf{Filtering}} & EntityLevel           & 71.04 / 19.56 & 76.31 / 12.69 & 91.10 / 72.08 & 94.19 / 64.42 & 65.13 / 23.32 & 20.67 / 20.29 \\
                                    & SummFC                & 72.73 / 19.98 & 74.62 / 12.15 & 91.71 / 72.16 & 94.79 / 65.12 & 65.86 / 22.02 & \textbf{23.20} / \textbf{22.57}\\
                                    & $\text{SummDSC}_{base}$       & \textbf{78.34} / 20.75 & 88.55 / 11.90 & 94.55 / 73.70 & 97.39 / 67.63 & 79.93 / \textbf{26.43} & 18.46 / 16.91 \\

\hline
\multirow{6}{*}{\textbf{Correction}} & SpanFact (our imple.) & 65.49 / 17.18 & 71.64 / 9.57  & 86.80 / 66.58 & 90.45 / 58.15 & 57.27 / 22.00 & 19.46 / 19.01 \\
                                     & BART-FC (our imple.)  & 65.28 / 20.28 & 68.76 / 13.07 & 86.99 / 70.22 & 92.20 / 62.23 & 61.11 / 24.81 & 18.60 / 19.25 \\
                                     & CogComp              & 69.76 / 18.59 & 74.40 / 9.21  & 90.23 / 70.14 & 93.25 / 60.77 & 62.97 / 22.85 & 20.19 / 19.72 \\
                                     & FactEdit              & 60.62 / 14.94 & 68.16 / 14.69 & 87.45 / 64.86 & 92.74 / 57.42 & 61.63 / 23.95 & 17.71 / 18.22 \\
                                     & CompEdit              & 70.31 / 18.49 & 72.61 / 9.31  & 90.60 / 70.08 & 94.13 / 61.25 & 63.83 / 23.21 & 19.13 / 20.25 \\
                                     & FactCloze (ours)      & 72.92 / 19.44 & 79.20 / 13.81 & 92.62 / 73.31 & 94.79 / 62.84 & 67.45 / 23.74 & 20.09 / 19.01 \\
\hline
\textbf{Combination} & $\text{SummDSC}_{base}$ + FactCloze      & 74.06 / \textbf{21.41} & \textbf{92.25} / \textbf{16.87} & \textbf{95.44} / \textbf{75.83} & \textbf{97.78} / \textbf{67.92} & \textbf{80.34} / 25.97 & 18.42 / 16.85 \\
\hline
\end{tabular}}
\cprotect\caption{Overall results on BART-generated summaries, where the left of the cell (*/*) indicates the CNN/DM while the right is XSum. \textbf{Correction} refers to the baselines of the factual error correction, which are applied to the BART-generated summaries. \textbf{Filtering} represents a BART that have been trained on datasets using various filtering methods. \textbf{Combination} briefly shows the integration of the two methods of filtering methods and factual error correction. The best performance is marked in bold. } 
\label{tab:3}
\end{table*}

\subsection{Performance on BART}

\subsubsection{$\textbf{SummDSC}_{base}$ for Summarization}

We firstly compare the factual consistency of the summaries generated by the summarization models trained on different faithful datasets. We train several summarization models on different faithful datasets and evaluate them on the original test set with five factual consistency metrics and ROUGE-2 (F1). Since ROUGE-2 requires a golden summary which is risky of factual inconsistency, it is only used as a reference standard for informativeness. As shown in Table \ref{tab:3}, we observe that the summaries generated by the model trained on $\text{SummDSC}_{base}$ achieve large improvements in most factual consistency metrics. This demonstrates the effectiveness of our data distillation method, which provides strong support for the subsequent training of FactCloze. However, it performs worse on ROUGE-2 compared to other baselines. We believe this is because the strict and sentence-level filtering destroys the informativeness in golden summaries to some extent, causing the model trained on them to generate shorter, less abstractive summaries. 

\subsubsection{FactCloze Performance on BART}
\label{sec:FactCloze Performance on BART}
We use five baselines and FactCloze\footnote{Based on previous conclusions, we use FactCloze-T5 on CNN/DM and FactCloze-BART on XSUM.} to correct the BART-generated summaries and show the results in the \textbf{Correction} of Table \ref{tab:3}.
For Rouge-2, all of the models in the \textbf{Correction} have a slight decrease, which is consistent with the phenomenon demonstrated in previous work. 
Besides, our method outperforms most of the baseline models on the majority of factual consistency metrics while most of baselines are not stable and even a decrease compared to the uncorrected summaries. Upon comparison with the filtering approach, it becomes evident that the post-editing methods generally falls worse. We believe that this is due to task gap between training and evaluation. Filtering-based methods are better suited to producing factually consistent summary distributions because they aim to train a summarization model. Conversely, post-editing methods rely on system-generated summaries for their post-editing tasks. This means their performance is inherently capped by the quality of the summaries they are tasked to correct.

\subsubsection{Combination}
Inspired by \citet{chaudhury2022x}, we try to further combine filtering methods and post-editing methods. Since $\text{SummDSC}_{base}$ and FactCloze outperform other baselines, we use them for our combination experiment. FactCloze is applied to correct the summaries generated by summarization model which is trained on $\text{SummDSC}_{base}$. It can be noted that combination approach further improves the faithfulness in most of factual consistency metrics. 

\subsection{Post-Alert Effectiveness}
\label{sec:Post-Alert Effectiveness}



Due to the difficulty in evaluating the post-alert mechanism automatically and fairly, we adopt an indirect approach. First, we define the corrected summaries containing \verb|<unk>| to raise alerts and count these samples as $n$. The $n$ samples will be discarded directly because they are considered risky sentences (\S \ref{sec:Post Alert}). Meanwhile, we introduce two baselines to discard the same number of samples. One baseline randomly drops samples (\textbf{Random}) and the other baseline discards last $n$ samples based on the descending order of the factual consistency scores averaged over the five factual consistency metrics (\textbf{Metric-base}). 
As shown in Tables \ref{tab:4} and \ref{tab:5}, \textbf{Post Alert} performs best overall and even achieves results over \textbf{Metric-base} on FactCloze-T5 \footnote{\textbf{Random} and \textbf{Metric-base} only use $\text{SummDSC}_{base}$, while \textbf{Post Alert} adds an additional dataset $\text{SummDSC}_{alert}$ to empower post-alert mechanism.}. This result indicates that post-alert mechanism can accurately capture the summaries that cannot be improved through factual error correction. Moreover, the introduction of a post-alert mechanism can also improve the accuracy of the correction, as confirmed by the fact that \textbf{Post Alert} outperforms \textbf{Metric-base} across the Tables.

\begin{table}
\scalebox{0.78}{
\begin{tabular}{lccc}
\hline
 \textbf{Metrics} & \textbf{Random} & \textbf{Post Alert} & \textbf{Metric-base} \\
 \hline
\textbf{QAFactEval}  & 62.06 / 16.25 & 65.55 / 17.87  & \textbf{67.97} / \textbf{18.28} \\
\textbf{SummaC}      & 83.67 / 32.61 & \textbf{87.97} / \textbf{36.47} & 85.55 / 30.56 \\
\textbf{ClozE}       & 87.39 / 65.62 & 91.09 / \textbf{69.24} &  \textbf{91.39} / 67.39 \\
\textbf{DAE}         & 92.70 / 55.13 & 95.38 / \textbf{59.25} & \textbf{96.22} / 56.73 \\
\textbf{FactCC}      & 73.69 / 29.06 & 77.60 / \textbf{34.29} & \textbf{82.16} / 31.10 \\
\hline
\end{tabular}}
\caption{Results of post-alert for FactCloze-BART, where the left of the cell (*/*) indicates the CNN/DM while the right is XSum. }
\label{tab:4}
\end{table}

\begin{table}
\scalebox{0.78}{
\begin{tabular}{lccc}
\hline
 \textbf{Metrics} & \textbf{Random} & \textbf{Post Alert} & \textbf{Metric-base} \\
 \hline
\textbf{QAFactEval} & 64.50 / 8.37  & \textbf{65.91} / \textbf{10.66} & 65.07 / 9.16 \\
\textbf{SummaC}     & 85.35 / 26.65 & \textbf{87.42} / \textbf{34.66} & 85.49 / 21.98 \\
\textbf{ClozE}      & 90.32 / 63.94 & \textbf{91.62} / \textbf{67.22} & 90.58 / 66.56 \\
\textbf{DAE}        & 95.88 / 49.95 & \textbf{97.27} / \textbf{58.35} & 96.11 / 51.76 \\
\textbf{FactCC}     & 77.57 / 23.58 & 79.81 / \textbf{29.71} & \textbf{80.10} / 26.28 \\
\hline
\end{tabular}}
\caption{Results of post-alert for FactCloze-T5.}
\label{tab:5}
\end{table}

\subsection{Ablation Study}
\label{sec:Ablation Study}
We conduct the ablation study on the self-diagnosis mechanism and the training dataset SummDSC. As shown in Table \ref{tab:2}, we observe a slight improvement in factual consistency scores with the self-diagnosis mechanism compared to the vanilla FactCloze in most cases. However, there is also a drop in several cases. We believe this instability is caused by the gap between training and inference. During training, we randomly select several factual factors to mask, while during inference, we expect the model to identify the factual errors and correct them. But the random training process does not provide the model with a stable recognition capability. For training datasets, we have demonstrated that FactCloze achieves better performance when trained on $\text{SummDSC}_{base}$ compared to other training datasets in \S \ref{sec:SummDSC Performance}. The analysis in \S \ref{sec:Post-Alert Effectiveness} reveals that training on $\text{SummDSC}_{alert}$ not only provides FactCloze with post-alert capability, but also further enhances its error correction accuracy. This improvement may be attributed to the fact that $\text{SummDSC}_{alert}$ serves as negative samples in relation to $\text{SummDSC}_{base}$, making the training process akin to contrastive learning and thereby improving the fact-awareness of FactCloze.

\subsection{Human Evaluation}
\label{sec:Human Evaluation}
We randomly sampled 100 document-hypothesis pairs from FRANK-XSum for manual annotation evaluation. Then, each document and hypothesis pair is labelled by two master students from our lab to evaluate the accuracy of factual error correction. Of the 100 samples, only 2\% are free of factual errors, while 66\% could not be corrected to enhance their faithfulness. As shown in Table \ref{tab:6}, FactCloze-BART achieves the best performance on \textbf{Complete} and \textbf{Alert}, while FactCloze-T5 performs better on \textbf{Partial}. It is worth noting that most models are less accurate on \textbf{Complete} than on \textbf{Partial}. This is because a number of hypotheses cannot be corrected as they bear no relation to the documents. This highlights the importance of our proposed post-alert mechanism.

\begin{table}
\scalebox{0.76}{
\begin{tabular}{lcccc}
\hline
 \textbf{Model} & \textbf{Partial(\%)} & \textbf{Complete(\%)} & \textbf{Alert(\%)}\\
\hline
\textbf{SpanFact} (our imple.) & 5.26 & 1.05 & / \\
\textbf{BART-FC} (our imple.) & 3.16 & 5.26 & / \\
\textbf{CogComp}  & 4.21 & 0.00 & / \\
\textbf{FactEdit} & 30.52 & 16.84 & / \\
\textbf{CompEdit} & 9.47 & 5.26 & / \\
\hline
\textbf{\textbf{FactCloze-BART}} & 40.82 & \textbf{25.51} & \textbf{61.07} \\
\textbf{\textbf{FactCloze-T5}} & \textbf{43.88} & 18.37 & 58.27 \\
\hline
\end{tabular}}
\caption{Results of manual annotations on XSum. \textbf{Partial} refers to the correction of only some errors while \textbf{Complete} indicates all errors have been corrected. F1 scores are used to evaluate \textbf{Alert} while others are accuracy scores.}
\label{tab:6}
\end{table}

\section{Case Study}
\label{sec:Case Study}
We present several correction cases using FactCloze in Tables \ref{tab:9} and \ref{tab:10} in the Appendix. In examples 1 to 3, FactCloze successfully identified unrelated hypothesis sentences. However, FactCloze-BART sometimes rewrites the sentence due to the cloze task and the construction of training datasets. The unrestricted generation and high extractive training datasets cause FactCloze-BART to favor extracting sentence from the document rather than only filling the blanks when confused. In examples 4 to 7, both FactCloze-BART and FactCloze-T5 show the factual error-correction abilities for entities and noun phrases. They effectively leverage the relationship between context and previous factual factors to generate accurate results. Moreover, we also present an example of causal modeling in Table \ref{tab:7} in the Appendix. We input two hallucinations to the decoder to disturb the generation of FactCloze. \textbf{Disturbed} raises an alert while \textbf{Automatic} obtains the correct sentence, which shows the effect of the previous factual factors for the one ready to be corrected.

\section{Conclusion}
In this paper, we propose a post-editing method FactCloze based on the conditional-generation cloze task for factual error correction. We show that our model can better improve the factual consistency of the summary than existing post-editing methods. In addition, we also propose a data distillation method and release a highly faithful dataset SummDSC. It can not only be used to train FactCloze but also for other tasks like summarization. We hope our findings in the paper will provide insights into future work in this direction.

\section{Acknowledgements}
This work was supported by the National Natural Science Foundation of China (Grant No.62176024).

\nocite{*}
\section{Bibliographical References}\label{sec:reference}


\newpage
\appendix
\section{Threshold Selection For SummDSC}
\label{sec:Threshold Selection}

As mentioned in \S \ref{sec:SummDSC}, we use DAE, ClozE and SummaC to select the faithful summaries and use ROUGE-2 precision (\textbf{R-2 Pre}) to generate the samples containing \verb|<unk>|s. We set different thresholds $\alpha_{DAE}$, $\alpha_{Summa}$, $\alpha_{ClozE}$ and $\alpha_{ROUGE}$ based on the values in Table \ref{tab:7}.
For CNN/DM, we set $\alpha_{DAE}=0.70$, $\alpha_{Summa}=0.45$, $\alpha_{ClozE}=0.70$ and $\alpha_{ROUGE}=0.30$. For XSum, we set $\alpha_{DAE}=0.50$, $\alpha_{Summa}=0.02$, $\alpha_{ClozE}=0.60$ and $\alpha_{ROUGE}=0.15$. We keep the samples that achieve higher scores than the thresholds on all three factual consistency metrics. And for the dropped samples, we keep the samples that gain lower scores than the threshold on ROUGE-2 precision and use them to generate \verb|<unk>|s. Following the filtering process, \textbf{SummDSC} retained 27.03\% of CNN/DM and 19.51\% of XSum, respectively.

\begin{table}[t]
\renewcommand\arraystretch{1.1}
\centering
\scalebox{0.75}{
\begin{tabular}{ll}
\hline
\multirow{4}{*}{\textbf{Document}} & (...) Temperton died in London last week\\
                                  & at the age of 66 after "a brief aggressive\\
                                  & battle with cancer", Jon Platt of Warner/\\
                                  & Chappell music publishing said. (...)\\
\hline
\multirow{2}{*}{\textbf{Hypothesis}} & \textcolor{red}{Templeton Templeton}, one of the UK's most\\
                                     & famous \textcolor{red}{66}, has died at the age of \textcolor{red}{74}.\\
\hline
\multirow{2}{*}{\textbf{Automatic}} & \textcolor{LimeGreen}{Rod Temperton}, one of the UK's most famous \\ 
                                    & \textcolor{LimeGreen}{songwriters}, has died at the age of \textcolor{LimeGreen}{66} .\\
                                    \cline{2-2}
\multirow{2}{*}{\textbf{Disturbed}} & \uline{Chaka Khan}, one of the UK's most famous \\
                                    & \uline{vocalists}, <unk>, has died at <unk>.\\
\hline
\end{tabular}}
\caption{An example represents the causal modeling in \textbf{FactCloze}, where \textbf{Automatic} indicates free correction and \textbf{Disturbed} is artificially filling in the incorrect factual factors (\uline{underlined}). The \textcolor{LimeGreen}{green} words indicate correct factual factors while \textcolor{red}{red} indicates incorrect ones.}
\label{tab:7}
\end{table}

\begin{table}
\centering
\scalebox{0.85}{
\begin{tabular}{lllll}
\hline
 & \textbf{R-2 Pre.} & \textbf{DAE} & \textbf{ClozE} & \textbf{SummaC}\\
\hline
\textbf{CNN/DM} & 0.4575 & 0.7155 & 0.6892 & 0.4595  \\
\textbf{XSum} & 0.1650 & 0.5392 & 0.6276 & 0.0675  \\
\hline
\end{tabular}}
\caption{Performance for summarization datasets on different metrics.}
\label{tab:8}
\end{table}

\begin{table*}
\begin{tabularx}{\textwidth}{lX}
\hline
\textbf{Document \#1} & (...) But while McHenry's reaction could very well have been a result of an overblown sense of entitlement, evidence of a mean girl who never left high school, what's also troubling is how quickly and gleefully the rest of us issued blame on McHenry without fully knowing -- or, it seems, caring about -- the other side of the story. The video that was released -- by the tow company -- was heavily edited and included only McHenry's responses, not the comments of the employee who may have provoked her and contributed to an argument that clearly escalates as the video goes on. McHenry knew she was being taped; (...)\\
\hline
\textbf{Summary} & The video was released on a video of her firing of a tow company .\\
\textbf{\textbf{FactCloze-BART}} & \texttt{<unk>} (\textbf{Alert})\\
\textbf{\textbf{FactCloze-T5}} & \texttt{<unk>} was released on \texttt{<unk>} of \texttt{<unk>} of \texttt{<unk>} . (\textbf{Alert})\\
\hline
\hline
\textbf{Document \#2} & (...) John Carver looks on as his Newcastle United struggle against rivals Sunderland in the Tyne-Wear derby Head coach John Carver said before the game he had a secret motivational tactic up his sleeve and would only reveal what it was following victory. Well, we'll never know what he used in a forlorn attempt to rouse his players. (...)\\
\hline
\textbf{Summary} & John Carver has been in a row for Newcastle united since the defeat .\\
\textbf{\textbf{FactCloze-BART}} & \texttt{<unk>} (\textbf{Alert})\\
\textbf{\textbf{FactCloze-T5}} & \texttt{<unk>} has been in \texttt{<unk>} for \texttt{<unk>} since \texttt{<unk>} . (\textbf{Alert})\\
\hline
\hline

\textbf{Document \# 3} & (...) Ferrari's Sebastian Vettel came second despite colliding with team-mate Kimi Raikkonen on the first lap. The incident damaged both cars, with Raikkonen fighting back to fifth behind the Red Bulls of Daniil Kvyat (...)\\
\hline
\textbf{Summary} & Kimi Raikkonen headed a Ferrari one-two in final practice at German grand prix.\\
\textbf{\textbf{FactCloze-BART}} & "Mercedes F1 boss Toto Wolff said the front wing coverage off it was because the car Rosberg was pretty damaged, " said the 31-year-old afterwards .\\
\textbf{\textbf{FactCloze-T5}} & <unk> headed a <unk> <unk> in <unk> at the <unk> <unk> . (\textbf{Alert})\\
\hline

\hline

\textbf{Document \#4} & (...) Due to the fire that it has suffered, the Sorrento may sink in the position in which it finds itself," the Balearic Islands port authority said in a tweet (in Spanish). (...)\\
\hline
\textbf{Summary} & Dozens of people have been injured in a fire at a cargo ship in \textcolor{red}{Spain's Canary Islands}, officials say.\\
\textbf{\textbf{FactCloze-BART}} & Dozens of people have been injured in a fire at a passenger ship in \textcolor{LimeGreen}{Spain's Balearic Islands}, officials say .\\
\textbf{\textbf{FactCloze-T5}} & Hundreds of people have been injured in a fire at a ferry in \textcolor{LimeGreen}{Spain's Balearic Islands}, officials say .\\

\hline

\end{tabularx}
\caption{Partial examples for \textbf{FactCloze}. Both the documents and summaries are from the FRANK dataset. Where \textcolor{LimeGreen}{green} words indicate correct factual factors, \textcolor{red}{red} ones indicate incorrect factual factors and \textbf{Alert} indicates the summary which cannot be corrected and we will raise an alert for it.}
\label{tab:9}
\end{table*}

\begin{table*}
\begin{tabularx}{\textwidth}{lX}
\hline

\textbf{Document \#5} & (...) Temperton died in London last week at the age of 66 after "a brief aggressive battle with cancer", Jon Platt of Warner/Chappell music publishing said. (...) Producer and DJ Mark Ronson wrote: "So devastated to hear that Rod Temperton has passed away. A wonderful man \& one of my favourite songwriters ever. " (...)\\
\hline
\textbf{Summary} & \textcolor{red}{Templeton Templeton}, one of the UK's most famous \textcolor{red}{66}, has died at the age of \textcolor{red}{74}.\\
\textbf{\textbf{FactCloze-BART}} & \textcolor{LimeGreen}{Rod Temperton}, one of the UK's most famous \textcolor{LimeGreen}{songwriters}, has died at the age of \textcolor{LimeGreen}{66} .\\
\textbf{\textbf{FactCloze-T5}} & \textcolor{LimeGreen}{Rod Temperton}, one of the UK's most famous \textcolor{LimeGreen}{songwriters}, has died at the age of \textcolor{LimeGreen}{66} .\\

\hline
\hline

\textbf{Document \#6} & Children in P6 and P7 will learn how to cope with change under the Healthy Me programme developed by Northern Ireland charity, Action Mental Health. Its chief executive David Babington said it will help prepare pupils for the stresses of the transfer test and big changes in their educational life. Five schools took part in a pilot. (...)\\
\hline
\textbf{Summary} & A secondary school in Northern Ireland has launched a new programme to improve \textcolor{red}{the stresses of a secondary school} in Northern Ireland.\\
\textbf{\textbf{FactCloze-BART}} & A secondary school in County Armagh has launched a new initiative to improve \textcolor{LimeGreen}{the mental wellbeing of a secondary school pupil} in northern ire .\\
\textbf{\textbf{FactCloze-T5}} & A school in County Armagh has launched a programme to improve \textcolor{LimeGreen}{the resilience of pupils in the transition period} .\\
\hline
\hline

\textbf{Document \#7} & Visitors to the Hebridean Celtic Festival will be able to use an app to trigger online information from items such as signs and posters on the site. Videos and band interviews will be among the online material available to view on phones and tablets. HebCelt is taking place in Stornoway on the Isle of Lewis from 19 to 22 July. The Waterboys, Imelda May, Lucy Spraggan, Skerryvore, Peatbog Faeries and Dougie MacLean are among this year's acts. HebCelt director Caroline Maclennan said: "We are offering the new augmented reality experience as an extra feature to add to the enjoyment of visiting the festival this year." (...)\\
\hline
\textbf{Summary} & A new augmented reality experience is being launched in \textcolor{red}{the isle of isle of lewis} as part of the \textcolor{red}{new augmented reality headset.}\\
\textbf{\textbf{FactCloze-BART}} & An augmented reality experience is being launched in \textcolor{LimeGreen}{Stornoway} as part of the \textcolor{LimeGreen}{Hebridean Celtic Festival} .\\
\textbf{\textbf{FactCloze-T5}} & An augmented reality experience is being launched in \textcolor{LimeGreen}{Scotland} as part of the \textcolor{LimeGreen}{HebCelt Festival} .\\
\hline

\end{tabularx}

\caption{Continuation of Table \ref{tab:9}.}
\label{tab:10}
\end{table*}

\begin{figure*}
\includegraphics[width=\textwidth]{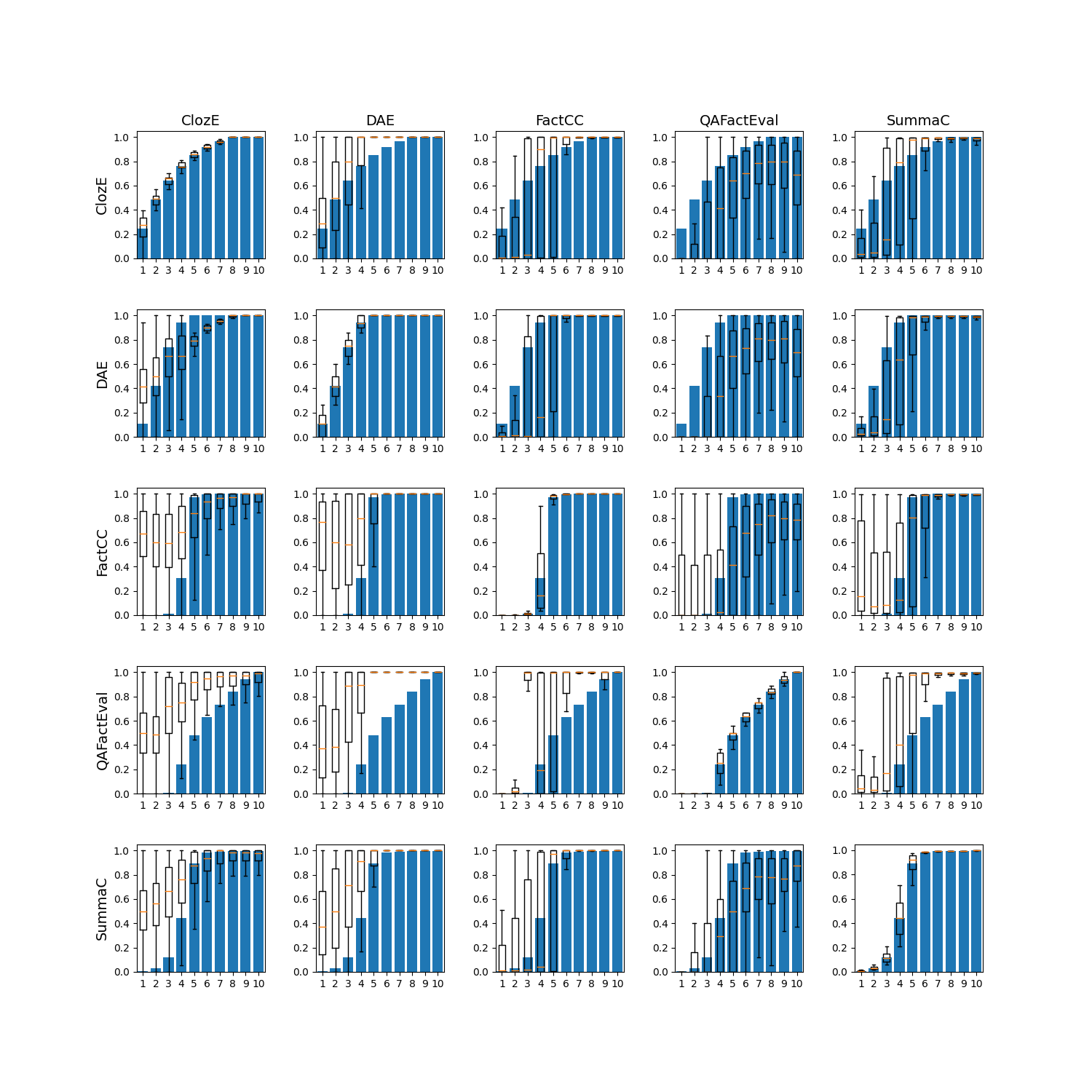}
\caption{A box plot for five factual consistency metrics. The samples are grouped into bins based on the percentiles of one metric score. The factual consistency score boxes of other metrics are plotted within each bin.}
\label{figure:4}
\end{figure*}

\end{document}